\documentclass{sig-alternate-2013}
\usepackage{hyperref}
\usepackage[ruled]{algorithm2e}

\newfont{\mycrnotice}{ptmr8t at 7pt}
\newfont{\myconfname}{ptmri8t at 7pt}
\let\crnotice\mycrnotice%
\let\confname\myconfname%

\permission{Permission to make digital or hard copies of all or part of this work for personal or classroom use is granted without fee provided that copies are not made or distributed for profit or commercial advantage and that copies bear this notice and the full citation on the first page. Copyrights for components of this work owned by others than ACM must be honored. Abstracting with credit is permitted. To copy otherwise, or republish, to post on servers or to redistribute to lists, requires prior specific permission and/or a fee. Request permissions from Permissions@acm.org.}
\conferenceinfo{MM'15,}{October 26--30, 2015, Brisbane, Australia. \\ 
{\mycrnotice{Copyright is held by the owner/author(s). Publication rights licensed to ACM.}}}
\copyrightetc{ACM \the\acmcopyr}
\crdata{978-1-4503-3459-4/15/10\ ...\$15.00.\\
DOI: http://dx.doi.org/10.1145/2733373.2806226 }

\begin{document}
%

\title{Temporal Localization of Fine-Grained Actions in Videos\\ by
Domain Transfer from Web Images}

\newcommand{\superscript}[1]{\ensuremath{^{\textrm{#1}}}}
\def\sharedaffiliation{\end{tabular}\newline\begin{tabular}{c}}
\def\wu{\superscript{*}}
\def\wg{\superscript{\dag}}

\numberofauthors{1} 
%
\author{
\alignauthor
\begin{tabular}{cccc}
Chen Sun\wu{\ } & Sanketh Shetty\wg{\ } & Rahul Sukthankar\wg{\ } & Ram Nevatia\wu
\end{tabular}
\sharedaffiliation
\begin{tabular}{ccc}
    \affaddr{{\wu}University of Southern California{\ }} & & \affaddr{{\wg}Google, Inc.{\ }} \\
    \email{\{chensun,nevatia\}@usc.edu} & & \email{\{sanketh,sukthankar\}@google.com}\\
  \end{tabular}
}
\maketitle
\begin{abstract}
  We address the problem of fine-grained action localization from temporally untrimmed web videos. We assume that only weak video-level annotations are available for training. The goal is to use these weak labels to identify temporal segments corresponding to the actions, and learn models that generalize to unconstrained web videos. We find that web images queried by action names serve as well-localized highlights for many actions, but are noisily labeled. To solve this problem, we propose a simple yet effective method that takes weak video labels and noisy image labels as input, and generates localized action frames as output. This is achieved by cross-domain transfer between video frames and web images, using pre-trained deep convolutional neural networks. We then use the localized action frames to train action recognition models with long short-term memory networks. We collect a fine-grained sports action data set FGA-240 of more than 130,000 YouTube videos. It has 240 fine-grained actions under 85 sports activities. Convincing results are shown on the FGA-240 data set, as well as the THUMOS 2014 localization data set with untrimmed training videos.
\end{abstract}




\section{Introduction}

This paper addresses the problem of fine-grained action localization
from unconstrained web videos. A fine-grained action takes place in a higher-level activity or event (e.g., \textit{jump shot} and \textit{slam dunk} in \textit{basketball}, \textit{blow candle} in \textit{birthday party}). Its instances are usually temporally localized within the videos, and share similar context with other fine-grained actions belonging to the same activity or event. 

Most existing work on action recognition focuses on action classification using pre-segmented short video clips~\cite{UCF101,Kuehne11,Schuldt2004_KTH}, which assumes implicitly that the actions of interest are temporally segmented during both training and testing. The TRECVID Multimedia Event Recounting
evaluation~\cite{MED13} as well as THUMOS 14 Challenge~\cite{THUMOS14}
both address action localization in untrimmed video, but the typical approach involves training classifiers on temporally segmented action clips and testing using sliding window on untrimmed video. This setting does not scale to large
action vocabularies, when data is collected from consumer video
websites.  Videos here are unconstrained in length, format (home
videos vs. professional videos), and almost always only have video
level annotations of actions.

\begin{figure}[t]
\begin{center}
\begin{tabular}{c}
\includegraphics[width=1.0\linewidth]{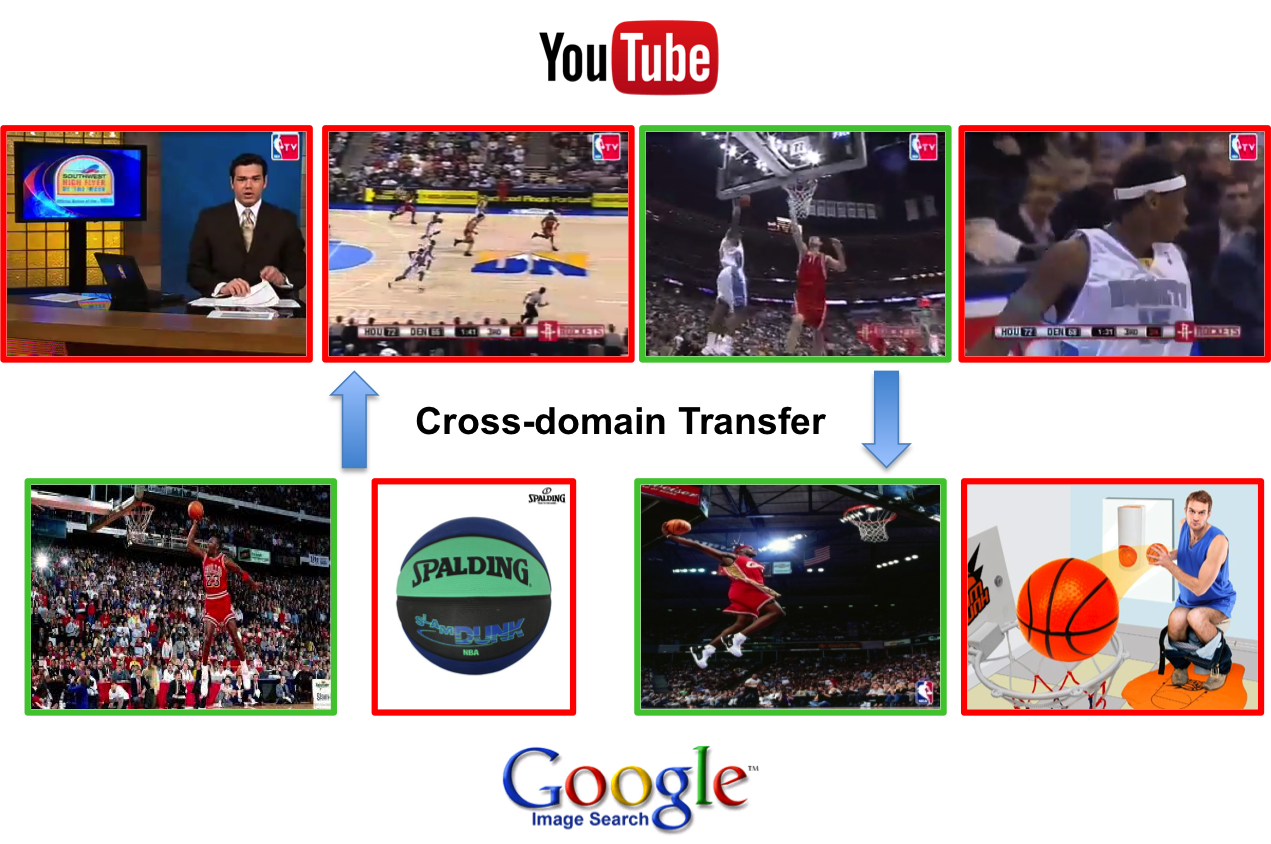}\\
\end{tabular}
\end{center}
\caption{Fine-grained actions are usually present as a tiny fraction
  within videos (top). Our framework uses cross-domain transfer from
  possibly noisy image search results (bottom) and identifies the action
  related images for both domains (marked in green).}
\label{fig:teaser}
\end{figure}

We assume that only video-level annotations are available for the
fine-grained action localization problem. The ability to localize fine
grained actions in videos has important applications
such as video highlighting, summarization, and automatic video
transcription. It is also a challenging problem for several
reasons: first, fine-grained actions for any high-level activity or event are inherently similar since they take place in similar scene context; second, occurrences of the fine-grained actions are usually short (a few seconds) in training videos, making it difficult to associate the video-level labels to the occurrences.


Our key observation is that one can exploit web images to help
localize fine-grained actions in videos. As illustrated in
Figure~\ref{fig:teaser}, by using action names (\textit{basketball
  slam dunk}) as queries, many of the image search results offer well
localized actions, though some of them are non-video like or
irrelevant. Identifying action related frames from weakly supervised
videos and filtering irrelevant image tags is hard in either modality
by itself; however, it is easier to tackle these two problems
together. This is due to our observation that although most of the video
frames and web images which correspond to actions are visually
similar, the distributions of non-action images from the video domain
and the web image domain are usually very different. For
example, in a video with a \textit{basketball slam dunk}, non slam
dunk frames in the video are mostly from a basketball game. The
irrelevant results returned by image search are more likely to be
product shots, or cartoons.

This motivates us to formulate a domain transfer problem between web
images and videos. To allow domain transfer, we first treat the videos as a bag of frames, and use the feature activations from deep convolutional neural networks (CNN)~\cite{AlexNet} as the common representation for images and
frames. Suppose we have selected a set of video frames and a set of web images for every action, the domain transfer framework goes in two directions: video frames to web images, and vice versa. For both directions, we use the selected images from the source domain to train action classifiers by fine-tuning the top layers of the CNN; we then apply the trained classifiers to the target domain. Each image in the target domain is assigned a confidence score given by its associated action classifier from the source domain. By gradually filtering out the images with low scores, the bidirectional domain transfer can progress iteratively. In practice, we start from the video frames to web images direction, and randomly select the video frames for training. Since the non-action related frames are not likely to occur in web images, the tuned CNN can be used to filter out the non-video like and irrelevant web images. The final domain transfer from web images is used to localize action related frames in videos. We term these action-related frames as \emph{localized action frames} (LAF).


Videos are more than an unordered collection of frames. We choose long
short-term memory (LSTM)~\cite{Hochreiter:1997:LSM:1246443.1246450}
networks as the temporal model. Compared with the traditional recurrent neural networks (RNN), LSTM has built-in \textit{input gates} and \textit{forget gates} to control its memory cells. These gates allow LSTM to either keep a long term memory or forget its history. The ability to learn from long sequences with unknown size of background is well-suited for fine-grained action localization from unconstrained web videos. 
We treat every sampled video frame as a time step in LSTM. When we train LSTM
models, we label all video frames by their video-level annotation, but
use the LAF scores generated by bidirectional domain transfer as weights on the loss for misclassification. By doing this, irrelevant frames are effectively down-weighted in the training stage. The framework can be naturally extended to use video shots as time steps, from which spatio-temporal features can be extracted to capture local motion information.



Fine-grained action localization from untrimmed web videos is a new task. 
The closest existing data set is THUMOS 2014 with 20 sports categories. It is designed for action localization using segmented videos as training, but has 1,010 untrimmed validation videos. To evaluate the framework in a large scale setting, we collected a new data set from YouTube. We chose 240 fine-grained actions belonging to 85 different sports activities, the total number of videos is over 130,000. Although the evaluated categories are sports actions, this method can be easily extended to other domains. For example, one can easily get \textit{cut cake, eat cake}
and \textit{blow candle} images for a \textit{birthday party} event
with image search.

Our work makes three major contributions:
\begin{itemize}
\item We show that learning temporally localized actions from videos
  becomes much easier if we combine weakly labeled video frames and
  noisily tagged web images. This is achieved by a simple yet
  effective domain transfer algorithm.

\item We propose a localization framework that uses LSTM network with the localized action frames to model the temporal evolution of actions.
\item We introduce the problem of fine-grained action localization with untrimmed videos, and collect a large fine-grained sports action data set with over 130,000 videos in 240 categories. The data set is available online.\footnote{\url{https://sites.google.com/site/finegrainedactions/}}
\end{itemize}

\begin{figure*}[t]
\begin{center}
\includegraphics[width=0.95\linewidth]{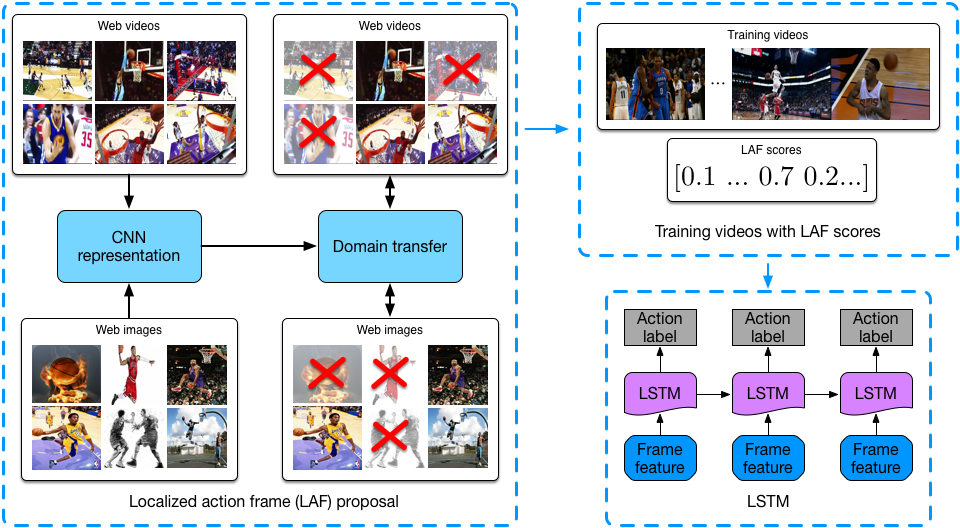}
\end{center}
\caption{Illustration of the LAF proposal framework for basketball slam dunk videos. We use a
  cross-domain transfer algorithm to jointly filter out non-video like web images and non-action like video frames. We then learn a
  LAF proposal model using the filtered images, which assigns LAF scores to training video frames. Finally, we train LSTM based fine-grained action detectors, where the misclassification penalty of each time step is weighted by the LAF score.}
\label{fig:kp_flow}
\end{figure*}

\section{Related Work}


Most existing work on activity recognition focuses on classifying
pre-segmented clips. For example, UCF 101 data set~\cite{UCF101} and
HMDB 51 data set~\cite{Kuehne11} provide 101 and 51 activity categories
respectively. Activity types range from primitive human actions,
sports to playing instruments; the typical length of each video clip
is 5 to 10 seconds. More recently, Karpathy et
al.~\cite{KarpathyCVPR14} proposed the Sports-1M data set with more
than 1 million untrimmed YouTube videos. Even though it offers 487
sports categories, most of them are high-level activities such as
\textit{basketball} and \textit{cricket}. For fine-grained action
recognition, Rohrbach et al.~\cite{rohrbach12cvpr} collected a cooking
action data set with temporal annotation, the videos were shot in an
indoor kitchen with static camera. To the best of our knowledge, there
is no previous work on fine-grained action localization with untrimmed
training videos.

Action recognition typically involves two basic steps: feature
extraction and classifier training. The standard approach is to
extract hand-designed low-level features, and then aggregate the
features into fixed-length feature vectors for classification. Oneata
et al.~\cite{FVeval_ICCV13} showed that a combination of visual
features (SIFT~\cite{Lowe04SIFT}) and motion features
(DT~\cite{Wang13DT}) represented using Fisher
Vectors~\cite{Perronnin07FV} produced state-of-the-art activity and
event classification performance.

Recent approaches, particularly those based on deep neural networks,
jointly learn features and classifiers. Karpathy et
al.~\cite{KarpathyCVPR14} proposed several variations of convolutional
neural net (CNN) architectures that extended Krizhevsky et al.'s image
classification model~\cite{AlexNet} and attempted to learn motion
patterns from spatio-temporal video patches. Simonyan and
Zisserman~\cite{DBLP:conf/nips/SimonyanZ14} obtained good results
on action recognition using a two-stream CNN that takes pre-computed
optical flow as well as raw frame-level pixels as input.

There have been many attempts to address the action localization problem. Tian et al.~\cite{TianSS13} proposed a temporal extension of the deformable part model (DPM) for action detection, they used spatio-temporal bounding boxes for training. Wang et al.~\cite{DBLP:conf/eccv/WangQT14} used dynamic poselets which took motion and pose into account. Jain et al.~\cite{DBLP:conf/cvpr/JainGJBS14} localized the actions with tubelets. All these approaches require manually annotated spatio-temporal bounding boxes for training. For temporal localization, THUMOS 2014~\cite{THUMOS14} data set provides trimmed video segments to train action classifiers, the classifiers can be used for localization with temporal sliding windows. 

The idea of using images as auxiliary data has also recently been explored. The most common usage is to learn mid-level
\textit{concept} detectors for high-level event
classification~\cite{DBLP:conf/mir/ChenCYLC14, HabibianICMR13,
  ChenICMR}. Yang et al.~\cite{Yang:2013:ETT:2457450.2457456} propose
a domain adaptation algorithm from images to videos, but assume
perfect image annotation. Divvala et
al.~\cite{DBLP:conf/cvpr/DivvalaFG14} learn mixtures of object
sub-categories using web images, they achieving this goal by
filtering sub-categories with low classification performance. 


The long short-term memory network (LSTM) was proposed by Hochreiter
and Schmidhuber~\cite{Hochreiter:1997:LSM:1246443.1246450} as an
improvement over traditional recurrent neural networks (RNN) for
classification and prediction of time series data.  Specifically, an
LSTM can \textit{remember} and \textit{forget} values from the past,
unlike a regular RNN where error gradients decay exponentially quickly
with the time lag between events. It has recently shown excellent
performance in modeling sequential data such as speech
recognition~\cite{DBLP:journals/corr/SakSB14,
  DBLP:journals/corr/abs-1303-5778}, handwriting
recognition~\cite{DBLP:conf/nips/GravesS08} and machine translation~\cite{DBLP:conf/nips/SutskeverVL14}. More recently, LSTM has also been applied to video-level action classification~\cite{DBLP:journals/corr/DonahueHGRVSD14,DBLP:journals/corr/SrivastavaMS15,Ng_2015_CVPR} and generate image descriptions~\cite{Vinyals_2015_CVPR,DBLP:journals/corr/KirosSZ14}.


\section{Our Approach}
Our proposed fine-grained action localization framework uses both weakly
labeled videos and noisily tagged web images. It employs the same CNN based representation for web images and video frameworks, and uses a bidirectional domain transfer algorithm to filter out irrelevant images in both domains. A localized action frame (LAF) proposal model is trained from the remaining web images, and used to assign LAF scores to video frames. Finally, we use long short-term memory networks to train fine-grained action detectors, using the LAF scores as the weight of loss for misclassification. The pipeline is illustrated in Figure~\ref{fig:kp_flow}.

\subsection{Shared CNN Representation}

A shared feature space is required for domain transfer between images and videos. Here we treat a video as a bag of frames, and extract activations from the intermediate layers of a convolutional neural network (CNN) as features for both web images and video frames. Although there is previous work on action recognition from still images using other representations~\cite{DBLP:journals/pami/YaoL12}, we choose CNN activations for its simplicity and state-of-the-art performance in several action recognition tasks~\cite{DBLP:conf/nips/SimonyanZ14, KarpathyCVPR14}.

Training a CNN end-to-end from scratch is time consuming, and requires a large amount of annotated data. It has been shown that CNN weights trained from large image data sets like ImageNet~\cite{imagenet_cvpr09} are generic, and can be applied to other image classification tasks by fine-tuning. It is also possible to disable the error back-propagation for the first several layers during fine-tuning. This is equivalent to training a shallower neural network using the intermediate CNN activations as features.

In this paper, we adopt the methodology of fine-tuning the top layers of CNN, and experiment with the AlexNet~\cite{AlexNet} CNN architecture. It contains five convolution layers and three fully connected layers. Each convolution layer is followed by a ReLU non-linearity layer and a maximum pooling layer. We pre-trained the network on ImageNet data set using the data partitions defined in~\cite{ILSVRCarxiv14}. We resized the images to 256 by 256, and used the raw pixels as inputs. For the purpose of fine-tuning, we fixed the network weights before its first fully connected layer and only updated the parameters of the top three layers. Feature activations from fc6 serve as the shared representation for web images and video frames, and allow cross-domain transfer between the two.



\begin{figure}[t]
\begin{center}
\begin{tabular}{c}
\includegraphics[width=0.96\linewidth]{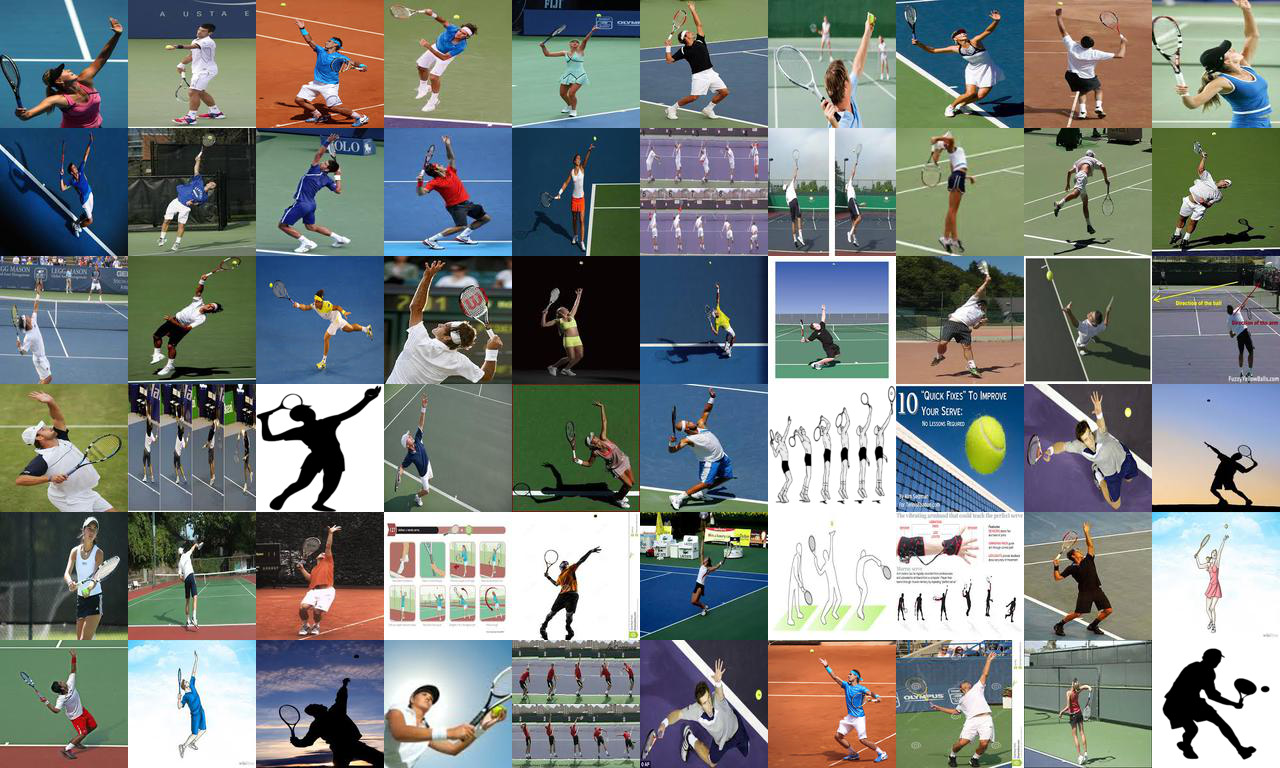}\\
\includegraphics[width=0.96\linewidth]{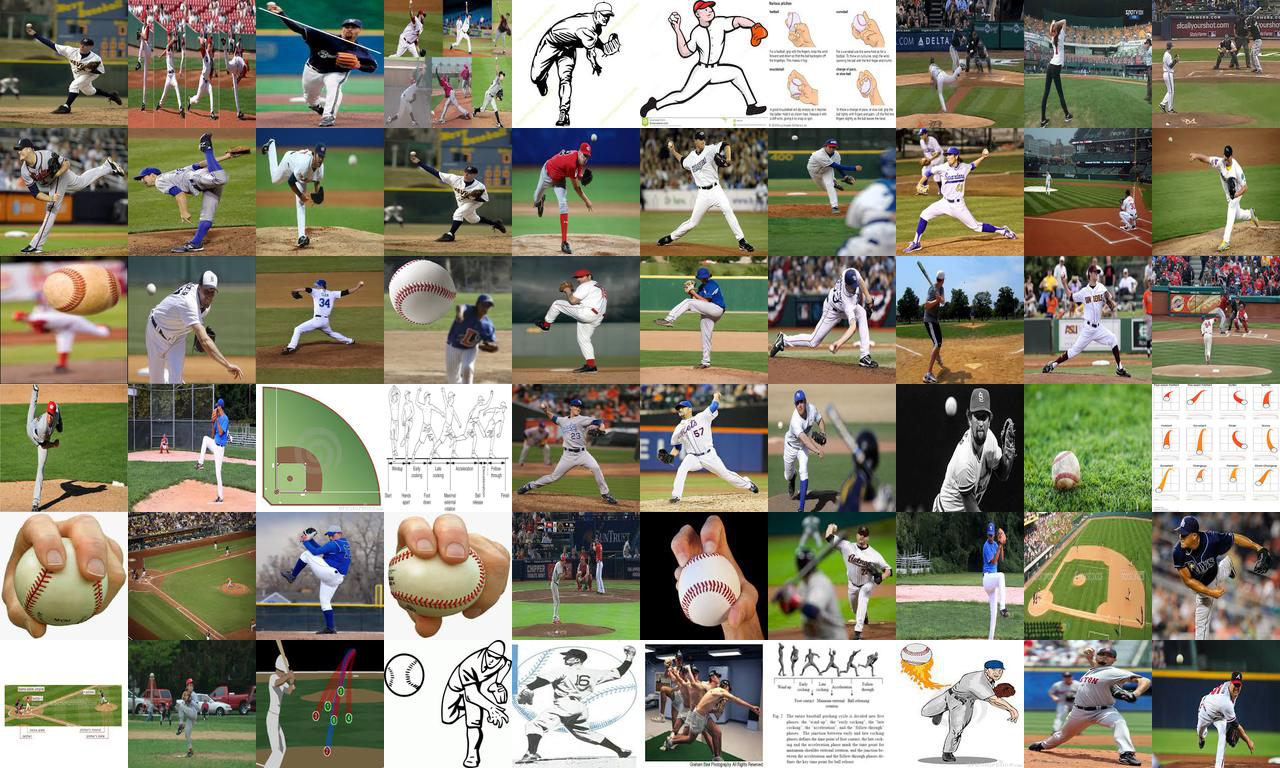}
\end{tabular}
\end{center}
\caption{Top retrieved images from the Google image search engine, using keywords \textit{tennis serve} and \textit{baseball pitch}.}
\label{fig:sample_images}
\end{figure}

\subsection{LAF Proposal with Web Images}

Fine-grained actions tend to be more localized in videos than
high-level activities. For example, a \textit{basketball} match video
usually consists of \textit{jump shot}, \textit{slam dunk},
\textit{free throw} etc., each of which may be as short as a few
seconds. We address the problem of automatically identifying them from minutes-long videos.



\begin{algorithm}
\SetKwInOut{Input}{Input}
\SetKwInOut{Output}{Output}
\Input{Images with noisy labels ${(I_i, a_i)}$, frames with video-level labels ${(V_i, a_i)}$}
\Output{LAF proposal model}
 Initialize $\mathcal{I}$ and $\mathcal{V}$ to include all image and frame inputs respectively.\\
 \While{stopping criteria not met}{
  1. Fine-tune $\textrm{CNN}_v$ using data in frame set $\mathcal{V}$.\\
  2. Compute $\textrm{CNN}_v(I)$ for all $I\in \mathcal{I}$.\\
  3. Update $\mathcal{I} = \{I|I\in\mathcal{I}, \textrm{CNN}_v(I)_{a_I} > \theta_1\}$.\\
  4. Fine-tune $\textrm{CNN}_i$ using data in image set $\mathcal{I}$.\\
  5. Compute $\textrm{CNN}_i(V)$ for all $V\in \mathcal{V}$.\\
  6. Update $\mathcal{V} = \{V|V\in\mathcal{V}, \textrm{CNN}_i(V)_{a_V} > \theta_2\}$.\\
 }
 \Return{$\textrm{CNN}_i$}
 \caption{Domain transfer algorithm for localized action frame proposal.}
 \label{alg:laf}
\end{algorithm}

Fortunately, we observe that many of the fine-grained actions have
\textit{image highlights} on the Internet (Figure~\ref{fig:sample_images}). They are easily obtained by querying image search engines with action names. However, these images are noisily labeled, and not useful for learning LAF proposal models directly, as they contain:
\begin{itemize}
\item Irrelevant images due to image crawling error, for example, a
  \textit{jogging} image could be retrieved with the keyword
  \textit{soccer dribbling}.
\item Items related to the actions, such as objects and logos.
\item Images with the same action but from a different domain, such as
  advertisement images with clear background, or cartoons.
\end{itemize}





Filtering the irrelevant web images is a challenging problem by
itself. However, it can be turned into an easier problem by using
weakly-supervised videos. We hypothesize that applying a classifier,
learned on video frames, as a filter on the images removes many irrelevant
images and preserves most video-like image highlights. More formally, assume we have video frames in $\mathcal{V}$ and web images in $\mathcal{I}$, and each of them is assigned a fine-grained action label $a=0,1,...,N-1$. We first learn a multi-class classifier $\textrm{CNN}_v(\cdot) \in \mathbb{R}^N$ by fine-tuning the top layers of CNN using video frames. $\textrm{CNN}_v(\cdot)$ encodes action discriminative information from the videos' perspective; we apply it to all $I \in \mathcal{I}$, and update 

\begin{equation}
\mathcal{I} = \{I|I\in\mathcal{I}, \textrm{CNN}_v(I)_{a_I} > \theta_1\}
\end{equation}
where $\theta_1\in [0,1]$ is the threshold for minimum softmax output, and $\textrm{CNN}_v(I)_{a_I}$ corresponds to the $a_I$-th dimension of $\textrm{CNN}_v(I)$.

We then use the filtered $\mathcal{I}$ to fine-tune $\textrm{CNN}_i(\cdot) \in \mathbb{R}^N$, and update $\mathcal{V}$ in a similar manner:

\begin{equation}
\mathcal{V} = \{V|V\in\mathcal{V}, \textrm{CNN}_i(V)_{a_V} > \theta_2\}
\end{equation}

We iterate the process and update $\mathcal{V}$ and $\mathcal{I}$ until certain stopping criteria are met. The LAF proposal model $\textrm{CNN}_i(\cdot)$ is learned using the final web image set $\mathcal{I}$, the LAF score for a video frame $V$ with action label $a$ is given by
\begin{equation}
\textrm{LAF}(V) = \textrm{CNN}_i(V)_{a}
\end{equation}
The whole process is summarized in Algorithm~\ref{alg:laf}.

\textbf{Discussion}: We initialize the frame set $\mathcal{V}$ by random sampling. Even though many of the sampled frames do not correspond to the actions of interest, they can help filter out the non-video like web images such as cartoons, object photos and logos. In practice, the random sampling of video frames is adequate for this step since the mis-labeled frames rarely appear in the web image collection.

We set the stopping criteria to be: (1) video-level classification accuracy on a validation set starts to drop; or (2) a maximum number of iterations is reached. To be more efficient, we train one-vs-rest linear SVMs using frames in $\mathcal{V}$ after each iteration, and apply the classifiers to video frames in the validation set. We take the average of frame-level classifier responses to generate video-level responses, and use them to compute classification accuracy.



\subsection{Long Short-Term Memory Network}
\label{sec:lstm}


Long Short-term Memory (LSTM)~\cite{Hochreiter:1997:LSM:1246443.1246450} is a type of recurrent neural network (RNN) that solves the vanishing and exploding gradients problem of previous RNN architectures when trained using back-propagation. Standard LSTM architecture
includes an input layer, a recurrent LSTM layer and an output
layer. The recurrent LSTM layer has a set of memory cells, which are
used to store real-valued state information from previous
observations. This recurrent information flow, from previous
observations, is particularly useful for capturing temporal evolution in
videos, which we hypothesize is useful in distinguishing between
fine-grained sports activities. In addition, LSTM's memory cells are protected by input gates and forget gates, which allow it to maintain a long-term memory and reset its memory, respectively. We employ the modification to LSTMs proposed by Sak et al.~\cite{DBLP:journals/corr/SakSB14} to add a projection layer after the LSTM layer. This reduces the dimension of stored states in memory cells, and helps to make the training process faster. 


Let us denote the input sequence $\mathbf{X}$ as $\{x_1, x_2, ..., x_T\}$, where in our case each $x_t$ is a feature vector of a video frame with time stamp $t$. LSTM maps the input sequence into the output action responses $\mathbf{Y} = \{y_1, y_2, ..., y_T\}$ by:

\begin{align}
i_t &= \sigma(W_{ix}x_t+W_{ir}r_{t-1}+W_{ic}c_{t-1}+b_i)\\
f_t &= \sigma(W_{fx}x_t+W_{rf}r_{t-1}+W_{cf}c_{t-1}+b_f)\\
c_t &= f_t \odot c_{t-1} + i_t \odot g(W_{cx}x_t+W_{cr}r_{t-1}+b_c)\\
o_t &= \sigma(W_{ox}x_t+W_{or}r_{t-1}+W_{oc}c_t+b_o)\\
m_t &= o_t \odot h(c_t)\\
r_t &= W_{rm}m_t\\
y_t &= W_{yr}r_t+b_y
\end{align}

\begin{figure}
\begin{center}
\begin{tabular}{c}
\includegraphics[width=1.0\linewidth]{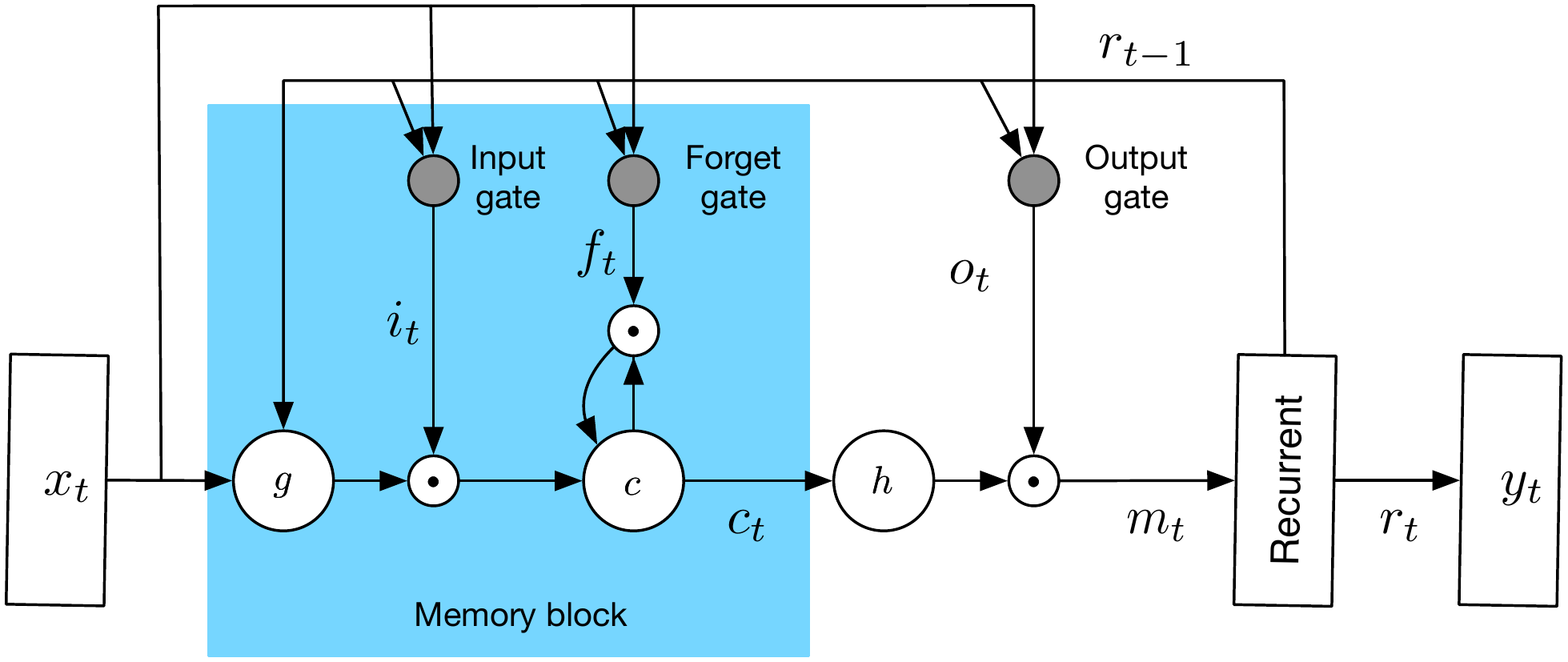}\\
\end{tabular}
\end{center}
\caption{Illustration of LSTM architecture with a single memory
  block. A recurrent projection layer is added to reduce the number of
  parameters. Reproduced from~\cite{DBLP:journals/corr/SakSB14} with the authors' permission.}
\label{fig:lstm_block}
\end{figure}

Here $W$'s and $b$'s are the weight matrices and biases, respectively, and $\odot$
denotes the element-wise multiplication operation. $c$ is the memory cell activation; $i, f, o$ are input gate, forget gate and output gate respectively. $m$ and $r$ are recurrent activation before and after projection. $\sigma$ is the sigmoid function, $g$ and $h$ are \emph{tanh}. An illustration of the LSTM architecture with a single memory block is shown in Figure~\ref{fig:lstm_block}.


\textbf{Training LSTM with LAF scores.} We sample video frames at 1
frame per second and treat each frame as a basic LSTM step. Similar to
speech recognition tasks, each time step requires a label and a
penalty weight for misclassification. The truncated backpropagation through time (BPTT) learning algorithm~\cite{Williams90anefficient} is used for training. We limit the maximum unrolling time steps to $k$ and only back-propagate the error for $k$ time steps. Incorporating the LAF scores into the LSTM framework is simple: we first run the LAF proposal pipeline to score all
sampled training video frames. Then we set the frame-level labels
based on video-level annotation, but use the LAF scores as the penalty
weights. Using this method, LSTM is \textit{forced} to make the
correct decision after watching a LAF returned by LAF proposal system,
and it is not penalized as heavily when gathering context information
from earlier frames or misclassifying an unrelated frame.

\textbf{Computing LAF scores for video shots.} For some data sets, it might be desirable to use video shots as the basic LSTM steps, as it allows the use of spatio-temporal motion features for representation. We extend the frame-level LAF scores to shot-level by taking the average of LAF scores from the sampled frames within a certain video shot.



\section{Experiments}
This section first describes the data set we collected for evaluation,
and then presents experimental results.

\subsection{Data Set}

\begin{figure}
\begin{center}
\begin{tabular}{c}
\includegraphics[width=1.0\linewidth]{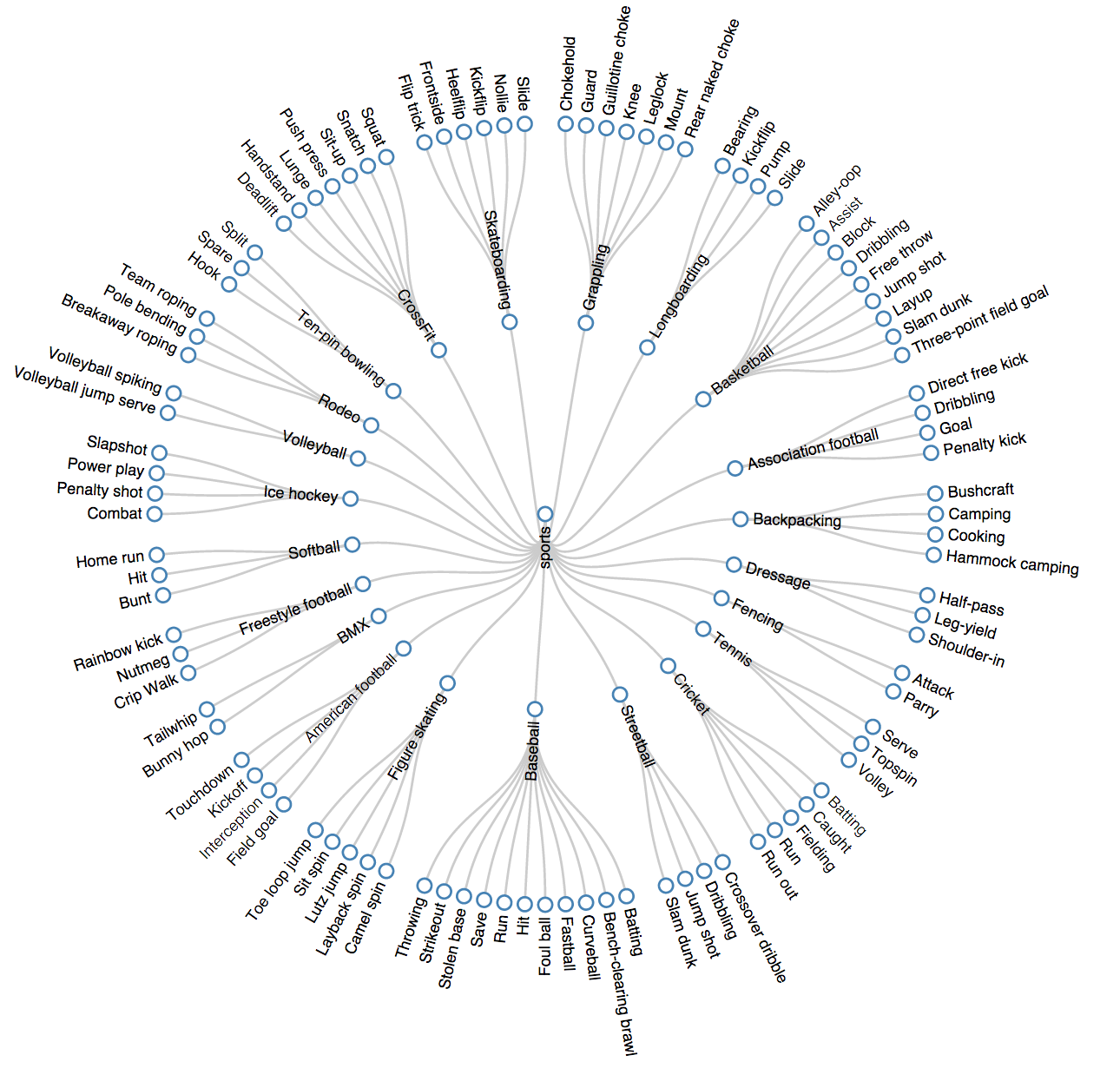}\\
\end{tabular}
\end{center}
\caption{Some of the high-level sports activities and their
  corresponding fine-grained sports actions in Fine Grained Actions
  240 data set. Best viewed under magnification.}
\label{fig:category_vis}
\end{figure}

There is no existing data set for fine-grained action
localization using untrimmed web videos. To evaluate our proposed method's
performance, we collected a Fine Grained Actions 240 (FGA-240) data set
focusing on sports videos. It consists of over 130,000 YouTube videos in 240
categories. A subset of the categories is shown in Figure
\ref{fig:category_vis}. We selected 85 high-level sports activities from the
Sports-1M data set~\cite{KarpathyCVPR14}, and manually chose the fine-grained actions take place in these activities. The action categories cover \textit{aquatic sports}, \textit{team sports}, \textit{sports with animals} and others.

We decided the fine-grained categories for each high-level sports
activity using the following method: given YouTube videos and their
associated text data such as titles and descriptions, we ran an
automatic text parser to recognize sports related entities. The
recognized entities which correlate with the high-level sports
activities were stored in the pool and then manually filtered to keep
only fine-grained sports actions. As an example, for
\textit{basketball} the initial entity pool contains not only
fine-grained sports actions (e.g., \textit{slam dunk},
\textit{block}), but also game events (e.g., \textit{NBA}) and
celebrities (e.g., \textit{Kobe Bryant}). Once the fine-grained
categories are fixed, we applied the same text analyzer to
automatically assign video-level annotations, and only kept the videos
with high annotation confidence. We finally visualized the data set to
filter out false annotations and removed the fine-grained sports
action categories with too few samples. 


Our final data set contains 48,381 training videos and 87,454
evaluation videos. The median number of training videos per category is 133. We used 20\% of the evaluation videos for
validation and 80\% for testing.

For temporal localization evaluation, we manually annotated 400 videos
from 45 fine-grained actions. The average length of the videos is
79 seconds.

\subsection{Experiment Setup}

\textbf{LSTM implementation.} We used the feature activations from pre-trained AlexNet (first fully-connected layer with 4,096 dimensions) as the input features for each time step. We followed the LSTM implementation by
Sak et al.~\cite{DBLP:journals/corr/SakSB14} which utilizes a multi-core
CPU on a single machine. For training, we used asynchronous stochastic
gradient descent and set batch size to 12. We tuned the training
parameters on the validation videos and set the number of LSTM cells
to 1024, learning rate to 0.0024 and learning rate decay with a factor
of 0.1. We fixed the maximum unroll time step $k$ to 20 to
forward-propagate the activations and backward-propagate the errors.

\textbf{Video level classification.} We evaluated fine-grained action
classification results on video level. We sampled test video frames at
1 frame per second. Given $T$ sampled frames from a video, these
frames are forward-propagated through time, and produce $T$ softmax
activations. We used average fusion to aggregate the frame-level activations over whole videos.

\textbf{Temporal localization.} We generated the frame-level softmax
activations using the same approach as video level classification. We
used a temporal sliding window of 10 time steps, the score of each
sliding window was decided by taking the average of softmax
activations. We then applied non-maximum suppression to remove the
localized windows which overlap with each other.

\textbf{Evaluation metric.} For classification, we report Hit @$k$, which is the percentage of testing videos whose labels can be found in the top $k$ results. For localization, we follow the same evaluation protocol
as THUMOS 2014~\cite{THUMOS14} and evaluate mean average precision. A
detection is considered to be a correct one if its overlap with
groundtruth is over some ratio $r$. 

\textbf{CNN baseline.} We deployed the single-frame architecture used by Karpathy et al.~\cite{KarpathyCVPR14} as the CNN baseline. It was shown to have comparable performance with multiple variations of CNNs while being simpler. We sampled the video frames at 1 frame per second, and used average fusion to aggregate softmax scores for classification and localization tasks.
Instead of training a CNN from scratch, we used network parameters from the pre-trained AlexNet, and fine-tuned the top two fully-connected layers and a softmax layer. Training parameters were decided using the validation set.

\textbf{Low-level feature baseline.} We extracted low-level features used by~\cite{KarpathyCVPR14, DBLP:conf/cvpr/YangT11} over whole videos for classification task, the feature set includes low-level visual and motion features aggregated using bag-of-words. We used the same neural network architecture as~\cite{KarpathyCVPR14} with multiple Rectified Linear Units to build classifiers based on the low-level features. Its structure
(e.g., number of layers, number of cells per layer) as well as training
parameters were decided with validation set.

\subsection{Video-level Classification Results}
We first report the fine-grained action classification performance on video
level.



\textbf{Comparison with baselines.} We compared several baseline
systems' performance against our proposed method on FGA-240 data set,
the results are shown in Table \ref{tab:acc}. From the table we can
see that systems based on CNN activations outperformed low-level
features by a large margin. There are two possible reasons for this:
first, CNN learned activations are more discriminative in classifying
fine-grained sports actions, even without capturing local motion
patterns explicitly; second, low-level features were aggregated on
video-level. These video-level features are more sensitive to background
and irrelevant video segments, which happens a lot in fine-grained sports
action videos.

Among the systems relying on CNN activations, applying LSTM gave
better performance than fine-tuning the top layers of CNN. While both
LSTM and CNN used the late fusion of frame-level softmax activations
to generate video-level classification results, LSTM took previous
observations into consideration with the help of memory cells. This
shows that temporal information helps classify fine-grained sports
actions, and it was captured by LSTM network.

\begin{table}
  \begin{center}
    \begin{tabular}{c|c|c}
      \hline
      Method & Video Hit @1 & Video Hit @5\tabularnewline
      \hline
      Random & 0.4 & 2.1\tabularnewline
      \hline
      Low-level features~\cite{DBLP:conf/cvpr/YangT11}& 30.8 & - \tabularnewline
      \hline
      CNN~\cite{KarpathyCVPR14} & 37.3 & 68.5 \tabularnewline
      \hline
      LSTM w/o LAF & 41.1 & 70.2 \tabularnewline
      LSTM w/ LAF & \textbf{43.4} & \textbf{74.9} \tabularnewline
      \hline
      \end{tabular}
  \end{center}
  \caption{Video-level classification performance of several different
    systems on fine-grained actions.}
  \label{tab:acc}
\end{table}

Finally, using LAF proposals helped further improve the video hit
@1 by 2.3\% and video hit @5 by 4.7\%. In Table
\ref{tab:lstm_ap_compare}, we show the relative difference in average
precision for LSTM with and without LAF proposal. We observe that
LAF proposal helps the most when the fine-grained sports
actions are likely to be identified based on single frames, and the
image highlights on the Internet are visually very similar to the
videos. Note that there are still non-video-like and irrelevant images
retrieved from the Internet for these categories, but the LAF
proposal system is an effective filter. Figure
\ref{fig:classification_examples} gives the three systems' output on a
few example videos.

\begin{figure}
\begin{center}
\includegraphics[width=0.95\linewidth]{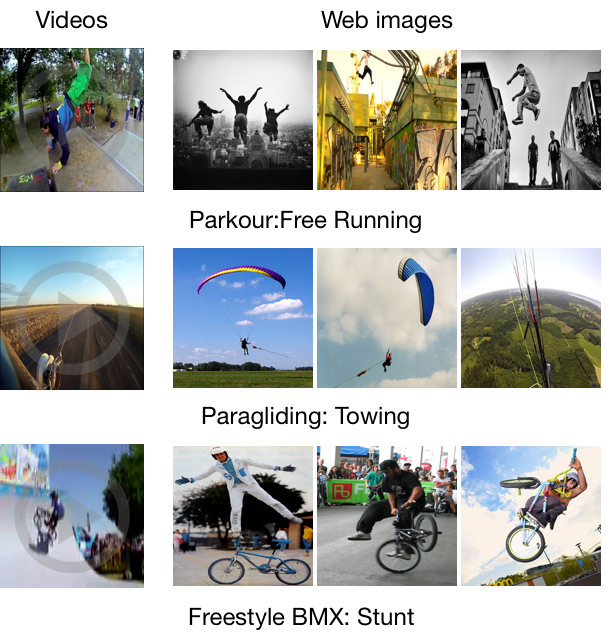}
\end{center}
\caption{Example actions when LAF is not helpful. The web images retrieved to generate LAF proposals might be beautified (top) or taken from different viewpoint (middle). Sometimes there is a mix of the two issues (bottom).}
\label{fig:bad_example}
\end{figure}

\begin{table}
\scriptsize
  \begin{center}
    \begin{tabular}{rr|ll}
      \hline
      Fine-grained sports & $\Delta$ AP & $\Delta$ AP & Fine-grained sports\tabularnewline
      \hline
      Fencing:Parry & 0.17 & -0.09 & Parkour:Free Running\tabularnewline
      Cricket:Run out & 0.15 & -0.08 & Freestyle soccer:Crip Walk\tabularnewline
      CrossFit:Deadlift & 0.10 & -0.08 & Paragliding:Towing\tabularnewline
      CrossFit:Handstand & 0.09 & -0.07 & Freestyle BMX:Stunt\tabularnewline
      Calisthenics:Push-up & 0.09 & -0.07 & Judo:Sweep\tabularnewline
      Rings:Pull-up & 0.08 & -0.06 & Basketball:Point \tabularnewline
      \hline
      \end{tabular}
  \end{center}
  \caption{Difference in average precision between LSTM
    with and without LAF proposal. Sorted by top wins (left) and top
    losses (right).}
  \label{tab:lstm_ap_compare}
\end{table}

We also identify several cases when LAF proposals failed to
work. The most common case is when most of the retrieved images are
non-video like but not filtered out. They could be posed images or
beautified images with logos, such as images retrieved for
\textit{Parkour:Free running}, or have different viewpoints than
videos, such as \textit{Paragliding:Towing}. Sample video snapshots and web images are shown in Figure~\ref{fig:bad_example}.


\begin{figure*}
\begin{center}
\includegraphics[width=0.95\linewidth]{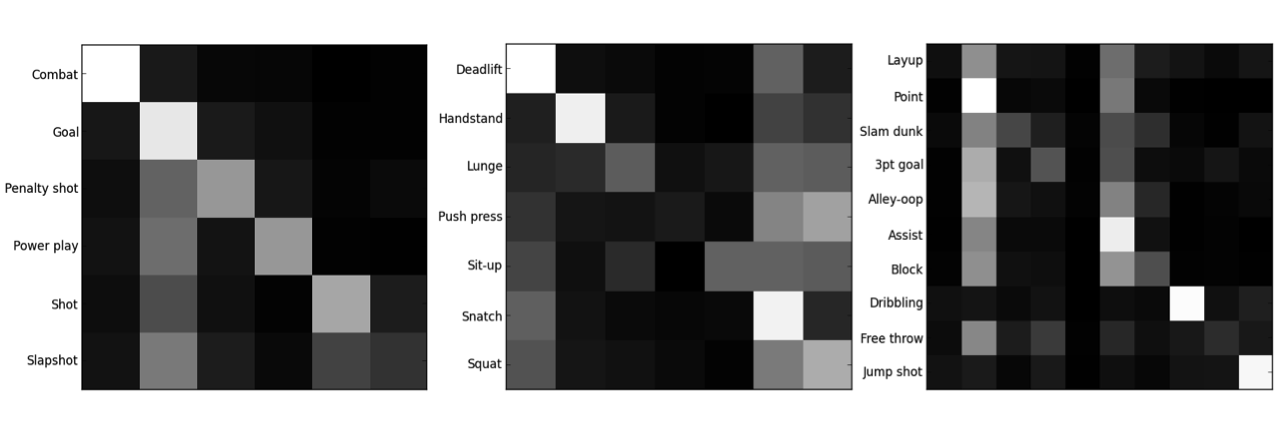}
\end{center}
\caption{Magnified view of confusion matrices for \textit{ice hockey},
  \textit{crossfit} and \textit{basketball}}
\label{fig:conf_mat}
\end{figure*}

\begin{figure*}
\begin{center}
\begin{tabular}{c}
\includegraphics[width=0.95\linewidth]{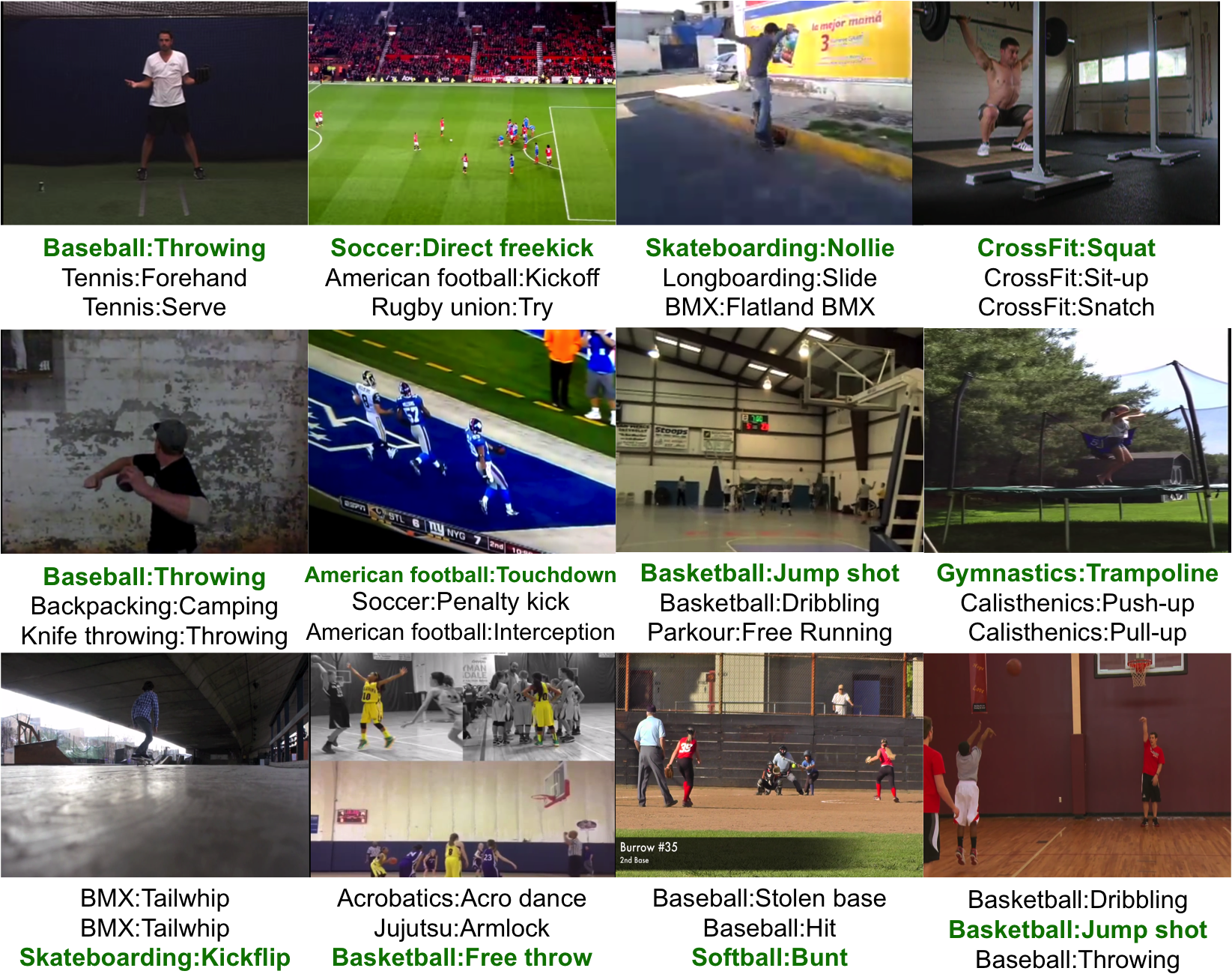}\\
\end{tabular}
\end{center}
\caption{Classification output for a few videos. The labels under each
video were generated by LSTM with LAF proposal, LSTM without
LAF proposal and CNN from top to
bottom. Correct answers are marked in bold.}
\label{fig:classification_examples}
\end{figure*}

\textbf{Impact of action hierarchy.} A fine-grained sports action
could be misclassified to either its sibling or non-sibling leaf nodes
in the sports hierarchy. For example, a \textit{basketball slam dunk}
can be confused with \textit{basketball alley-oop} as well as
\textit{street ball slam dunk}. To study the source of confusion, we
decided to measure classification accuracy of high-level sports
activities, and check how the numbers compared with fine-grained
sports actions.

We obtain the confidence values for high-level sports activities by
taking the average of their child nodes' confidence scores. Table
\ref{tab:hierarchy} shows the classification accuracy with different
methods. We can see that the overall trend is the same as fine-grained
sports actions: LSTM with LAF proposal is still the
best. However, the numbers are much higher than when measured on
fine-grained level, which indicates that the major source of confusion
still comes from the fine-grained level. In Figure \ref{fig:conf_mat},
we provide the zoom-in confusion matrices for \textit{ice hockey},
\textit{crossfit} and \textit{basketball}.

\begin{table}
  \begin{center}
    \begin{tabular}{c|c|c}
      \hline
      Method & Video Hit @1 & Video Hit @5\tabularnewline
      \hline
      Random & 1.2 & 5.9 \tabularnewline
      \hline
      CNN~\cite{KarpathyCVPR14}& 69.2 & 75.9 \tabularnewline
      \hline
      LSTM w/o LAF& 71.7 & 77.3 \tabularnewline
      LSTM w/ LAF& \textbf{73.6} & \textbf{79.5} \tabularnewline
      \hline
      \end{tabular}
  \end{center}
  \caption{Classification performance when measured on
    high-level sports activities (e.g., \textit{basketball, soccer}).}
  \label{tab:hierarchy}
\end{table}



\subsection{Localization Results}
\label{sec:local_result}

\begin{figure}[t]
\begin{center}
\begin{tabular}{c}
\includegraphics[width=1.0\linewidth]{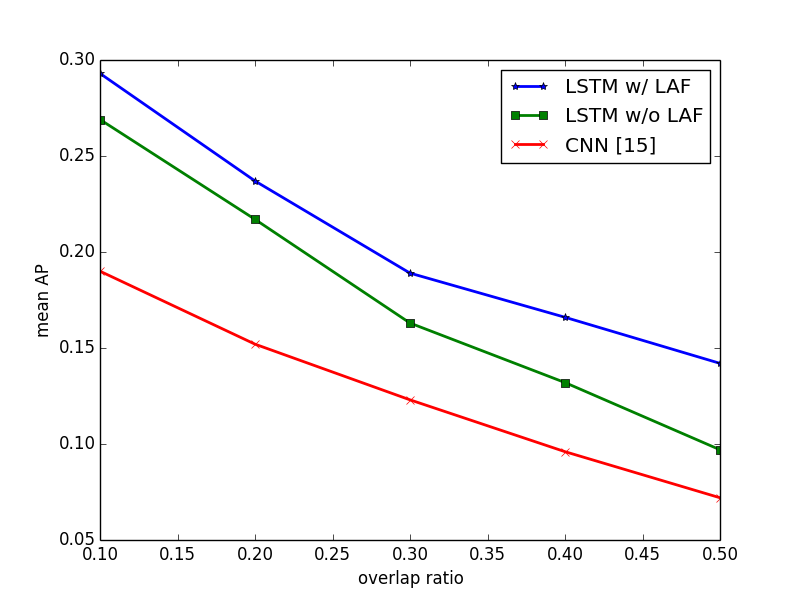}\\
\end{tabular}
\end{center}
\caption{Temporal localization performance on FGA-240 data set.}
\label{fig:detect_map}
\end{figure}

\textbf{Comparison with baselines.} We applied the frameworks to localize fine-grained actions, and varied the overlap ratio $r$ from 0.1 to 0.5 for evaluation. Figure~\ref{fig:detect_map} shows the mean average precision over all 45
categories of different systems. We did not include the baseline using
low-level features for evaluation as they were computed over whole
videos. From the figure we can see that LSTM with LAF proposal
outperformed both CNN and LSTM without LAF proposal significantly, the gap grows wider as we increase the overlap ratio. This
confirms that temporal information and LAF proposal are helpful
for the temporal localization task.

\begin{table}
\small
  \begin{center}
    \begin{tabular}{rr|ll}
      \hline
      Fine-grained sports & $\Delta$ AP & $\Delta$ AP & Fine-grained sports\tabularnewline
      \hline
      Soccer:Penalty kick & 0.32 & -0.06 & Baseball:Run\tabularnewline
      Tennis:Serve & 0.25 & -0.01 & Skateboarding:Kickflip\tabularnewline
     Basketball:Dribbling & 0.21 & -0.01 & Volleyball:Spiking\tabularnewline
      \hline
      \tabularnewline
      \hline
      Fine-grained sports & $\Delta$ AP & $\Delta$ AP & Fine-grained sports\tabularnewline
      \hline
      Baseball:Brawl & 0.52 & -0.11 & Streetball:Crossovers\tabularnewline
      Ice hockey:Combat & 0.48 & -0.05 & Ice hockey:Penalty shot\tabularnewline
      Soccer:Penalty kick & 0.33 & -0.04 & Fencing:Parry \tabularnewline
      \hline
      \end{tabular}
  \end{center}
  \caption{Difference in average precision, compared between LSTM with
  and without LAF proposal (top), LSTM with LAF proposal and
CNN (bottom), overlap ratio is fixed to 0.5. A positive number means
LSTM with LAF proposal is better.}
  \label{tab:detect_ap_diff}
\end{table}

In Table \ref{tab:detect_ap_diff}, we show the most different average
precisions on action level. Some actions have clearly benefited from
the introduction of LSTM as well as LAF proposal. We also observed
that some actions were completely missed by all three systems, such as
\textit{Baseball:Hit}, \textit{Basketball:Three-point field goal} and
\textit{Basketball:Block}, possibly due to the video frames
corresponding to these actions not being well localized during
training.

\subsection{Localization Results on THUMOS 2014}

To verify the effectiveness of domain transfer from web images, we also conducted a localization experiment on the THUMOS 2014 data set~\cite{THUMOS14}. This data set consists of over 13,000 temporally trimmed videos from 101 actions, 1,010 temporally untrimmed videos for validation and 2,574 temporally untrimmed videos for testing. The localization annotations cover 20 out of the 101 actions in the validation and test sets. All 20 actions are sports related.

\textbf{Experiment setup}: As this paper focuses on temporal localization of untrimmed videos, we dropped the 13,000 trimmed videos, and used the untrimmed validation videos as the only positive samples for training. We also used 2,500 background videos as the shared negative training data.

To generate LAF scores, we downloaded web images from Flickr and Google using the action names as queries. We also sampled training video frames at 1 frame per second. We used the AlexNet features for domain transfer experiment.

Recently, it has been shown that a combination of improved dense trajectory features~\cite{wang:hal-00873267} and Fisher vector encoding~\cite{Perronnin07FV} (iDT+FV) offers the state-of-the-art performance on this data set. This motivated us to switch LSTM time steps from frames to video segments, and represent segments with iDT+FV features for the final detector training. We segmented all videos uniformly with a window width of 100 frames and step size of 50 frames. For iDT+FV feature extraction, we took only the MBH modality with 192 dimensions and reduced the dimensions to 96 with PCA. We used the full Fisher vector formulation with the number of GMM cluster centers set to 128. The final video segment representation has 24,576 dimensions.

\textbf{Results}: We compared the performance of LSTM weighted by LAF scores against several baselines. \textbf{LSTM w/o LAF} randomly assigned misclassification penalty for each step of LSTM, where 30\% of the steps were set to 1, and others 0. The \textbf{Video} baseline used iDT+FV features aggregated over whole videos to train linear SVM classifiers, and applied the classifiers to the testing video shots. It was used by~\cite{ChenICMR,potapov2014category} and achieved state-of-the-art performance in event recounting and video summarization tasks. None of these systems require temporal annotations. Finally, \textbf{Ground truth} employed manually annotated temporal localizations to set LSTM penalty weights. It is used to study the performance difference between LAF and an oracle with perfect localized actions.

Table~\ref{tab:thumos_performance} shows the mean average precision for the four approaches. As expected, using manually annotated ground truth for training provides the best localization performance. Although LSTM with LAF scores has worse performance than using ground truth, it outperforms LSTM without LAF scores, and the video-level baseline by large margins. This further confirms that LAF proposal by domain transfer from web images is effective in action localization tasks.

\begin{table}
\small
  \begin{center}
    \begin{tabular}{c|ccccc}
      \hline
      &\multicolumn{5}{c}{Overlap ratio}\\
      \hline
      Method& 0.1 & 0.2 & 0.3 & 0.4 & 0.5\\
      \hline
      Ground truth & 0.161 & 0.152 & 0.112 & 0.071 & 0.044\\ 
      \hline
      Video~\cite{ChenICMR,potapov2014category} & 0.098 & 0.089 & 0.071 & 0.041 & 0.024\\
      LSTM w/o LAF & 0.076 & 0.071 & 0.057 & 0.038 & 0.024\\
      LSTM w/ LAF & \textbf{0.124} & \textbf{0.110} & \textbf{0.085} & \textbf{0.052} & \textbf{0.044}\\
      \hline
      \end{tabular}
  \end{center}
  \caption{Temporal localization on the test partition of THUMOS 2014 dataset. Ground truth uses temporal annotation of the training videos.}
  \label{tab:thumos_performance}
\end{table}

\section{Conclusion}
We studied the problem of fine-grained action localization for temporally untrimmed web videos. We proposed to use noisily tagged web images to discover
localized action frames (LAF) from videos, and model temporal
information with LSTM networks. We conducted thorough evaluations on
our collected FGA-240 data set and the public THUMOS 2014 data set, and showed the effectiveness of LAF proposal by domain transfer from web images.

\section{Acknowledgments}
We thank George Toderici, Matthew Hausknecht, Jia Deng, Weilong Yang, Susanna Ricco, Tomas Izo, Thomas Leung, Congcong Li and Howard Zhou for helpful comments and discussions. We also thank Bernard Ghanem, Fabian Caba Heilbron and Juan Carlos Niebles for kindly providing us video annotation tools.
%
\bibliographystyle{abbrv}
{\normalsize\bibliography{egbib}}  
\end{document}